# SD-CNN: a Shallow-Deep CNN for Improved Breast Cancer Diagnosis

Fei Gao, Teresa Wu, Jing Li, Bin Zheng, Lingxiang Ruan, Desheng Shang and Bhavika Patel

*Abstract*—Breast cancer is the second leading cause of cancer death among women worldwide. Nevertheless, it is also one of the most treatable malignances if detected early. Screening for breast cancer with full field digital mammography (FFDM) has been widely used. However, it demonstrates limited performance for women with dense breasts. An emerging technology in the field is contrast-enhanced digital mammography (CEDM), which includes a low energy (LE) image similar to FFDM, and a recombined image leveraging tumor neoangiogenesis similar to breast magnetic resonance imaging (MRI). CEDM has shown better diagnostic accuracy than FFDM. While promising, CEDM is not yet widely available across medical centers. In this research, we propose a Shallow-Deep Convolutional Neural Network (SD-CNN) where a shallow CNN is developed to derive "virtual" recombined images from LE images, and a deep CNN is employed to extract novel features from LE, recombined or "virtual" recombined images for ensemble models to classify the cases as benign vs. cancer. To evaluate the validity of our approach, we first develop a deep-CNN using 49 CEDM cases collected from Mayo Clinic to prove the contributions from recombined images for improved breast cancer diagnosis (0.85 in accuracy, 0.84 in AUC using LE imaging vs. 0.89 in accuracy, 0.91 in AUC using both LE and recombined imaging). We then develop a shallow-CNN using the same 49 CEDM cases to learn the nonlinear mapping from LE to recombined images. Next, we use 89 FFDM cases from INbreast, a public database to generate "virtual" recombined images. Using FFDM alone provides 0.84 in accuracy (AUC=0.87), whereas SD-CNN improves the diagnostic accuracy to 0.90 (AUC=0.92).

## 1. INTRODUCTION

Although about 1 in 8 U.S. women (~12%) will develop invasive breast cancer over the course of her lifetime (U.S. Breast Cancer Statistics, 2018), breast cancer death rates have been steadily and/or significantly decreasing since the implementation of the population-based breast cancer screening program in late 1970s due to the early cancer detection and the improved cancer treatment methods (Rosenquist and Lindfors, 1998). Among the existing imaging modalities, full field digital mammography (FFDM) is the only clinically acceptable imaging modality for the population-based breast cancer screening, while Ultrasound (US) and Magnetic Resonance Imaging (MRI) are also used as adjunct imaging modalities to mammography for certain special subgroups of women (Lehrer et al., 2012). However, using FFDM is not an optimal approach in breast cancer screening due to its relatively low detection sensitivity in many subgroups of women. For example, although FFDM screening has an overall cancer detection accuracy of 0.75 to 0.85 in the general population, its accuracy in several subgroups of the high-risk women including those with positive BRCA (BReast CAncer) mutation or dense breasts decreases to 0.30 to 0.50 (Elmore et al., 2005). On the other hand, using dynamic contrast enhanced breast MRI can yield significantly higher cancer detection performance due to its ability to detect tumor angiogenesis through contrast enhancement and exclude suspicious dense tissues(Warner et al., 2004). Yet, its substantially higher cost, lower accessibility and longer imaging scanning time forbids breast MRI being used as a primary imaging modality in breast cancer screening and detection. In addition, lower image resolution of breast MRI is a disadvantage as comparing to FFDM.

In order to combine the advantages of both FFDM and MRI, a new novel imaging modality namely,

contrast-enhanced digital mammography (CEDM) emerges and starts to attract broad research and clinical application interest. CEDM is a recent development of digital mammography using the intra-venous injection of an iodinated contrast agent in conjunction with a mammography examination. Two techniques have been developed to perform CEDM examinations: the temporal subtraction technique with acquisition of high-energy images before and after contrast medium injection and the dual energy technique with acquisition of a pair of low and high-energy images only after contrast medium injection. During the exam, a pair of low and high-energy images is obtained after the administration of a contrast medium agent. The two images are combined to enhance contrast uptake areas and the recombined image is then generated(Fallenberg et al., 2014). In CEMD, it has low energy (LE) imaging, which is comparable to routine FFDM and recombined imaging similar to breast MRI. Comparing to breast MRI, CEDM exam is about 4 times faster with only about 1/6 the cost (Patel et al., 2017). In addition, CEDM imaging has 10 times the spatial resolution of breast MRI. Therefore, CEDM can be used to more sensitively detect small residual foci of tumor, including calcified Ductal Carcinoma in Situ (DCIS), than using MRI (Patel et al., 2017). Several studies including prospective clinical trials conducted at Mayo Clinic have indicated that CEDM is a promising imaging modality that overcomes tissue overlapping ("masking") occurred in FFDM, provides tumor neovascularity related functional information similar to MRI, while maintaining high image resolution of FFDM (Cheung et al., 2014; Fallenberg et al., 2014; Gillman et al., 2014; Luczyńska et al., 2014). Unfortunately, CEDM as a new modality is yet widely available in many other medical centers or breast cancer screening facilities in the U.S. and/or across the world limiting its broad clinical impacts.

In clinical breast imaging (US, MRI, FFDM and CEDM), reading and interpreting the images remains a difficult task for radiologists. Currently, breast cancer screening has high false positive recall rate (i.e., $\geq 10\%$). Computer-aided detection (CADe) and diagnosis (CADx) schemes (Tan et al., 2014; Carneiro et al., 2017; Gao et al., 2016; Muramatsu et al., 2016) have been developed and demonstrated the clinical potentials to be used as "the second reader" to help improve radiologists' performance in the diagnosis. In order to overcome the limitation of lower accessibility to CEDM systems and help radiologists more accurately conduct the diagnosis, this research proposes the development and validation of a new CADx scheme, termed Shallow-Deep Convolutional Neural Network (SD-CNN). SD-CNN combines image processing and machine learning techniques to improve the malignancy diagnosis using FFDM by taking advantages of information available from the CEDM.

CNN is a feed-forward artificial neural network that has been successfully implemented in the broad computer vision areas for decades (Lecun et al., 2015; LeCun et al., 1998). As it evolves, different CNN models have been developed and implemented. The computational resource and devices available in recent years make the training of CNN with large number of layers (namely, the deep CNN) possible. Applying deep CNNs in image recognition was probably first demonstrated in ImageNet competition (Russakovsky et al., 2015) back in 2012. Since then, it has become a popular model for various applications ranging from natural language processing, image segmentation to medical imaging analysis (Cha et al., 2016; Tajbakhsh et al., 2016, Wang et al., 2017). The main power of a deep CNN lies in the tremendous trainable parameters in different layers (Eigen et al., 2013; Zeiler and Fergus, 2014). These are used to extract discriminative features at different level of abstraction (Tajbakhsh et al., 2016). However, training a deep CNN often requires a large volume of labeled training data, which may not be easily available in medical applications. Secondly, training a deep CNN requires massive computational resources, as well as rigorous research in architecture design and hyper-parameters tuning. To address these challenges, a promising solution is transfer learning (Banerjee et al. 2017), that is, a deep CNN model is trained followed by a task-specific parameter fine-tuning process. The trained models are established by experienced researchers using publicly labeled image datasets. For a specific task, the model is often treated as a feature generator to extract features describing the images from abstract level to detailed levels. One can then develop classification models (SVMs, ANNs, etc.) using the derived features. Promising results have been reported in several medical applications, such as chest pathology identification (Bar et al., 2015), breast mass detection and classification (Samala et al., 2016), just to name a few. While exciting, earlier CNN models such as AlexNet (Krizhevsky et al., 2012), GoogLeNet (Simonyan and Zisserman, 2014) and VGGNet (Szegedy et al., 2014) are known to suffer from gradient vanishing when the number of layers increases significantly. A newer model, ResNet (He et al., 2014) with a "short-cut" architecture is recently proposed to address the issue. The imaging competition results show the ResNet outperforms other CNN models by at least 44% in classification accuracy.

The potentials CNN brings to medical imaging research are not limited to deep CNN for imaging feature extraction. A second area that medical research can benefit is indeed using CNN for synthetic image rendering. Here an image is divided into a number of smaller patches fed into a CNN (e.g., 4-layer CNN in this research) as the input and the output is a synthetic image. The CNN is trained to learn the non-linear mapping between the input and output images. Several successful applications have been reported, such as synthesizing positron emission tomography (PET) imaging (Li et al., 2014) or CT image (Han, 2017; Nie et al., 2016) from MRI image, and from regular X-ray to bone-suppressed recombined X-ray (Yang et al., 2017).

Motivated by this two-fold applicability of CNN, this research proposes a Shallow-Deep CNN (SD-CNN) as a new CAD scheme to tackle the unique problem stemmed from the novel imaging modality, CEDM, for breast cancer diagnosis. Our first hypothesis is that applying a deep CNN to CEDM is capable of taking advantage of recombined imaging for improved breast lesion classification due to the contribution from the tumor functional image features. Second, in order to expand the advantages of CEDM imaging modality to the regular FFDM modality, we hypothesize that a shallow CNN is capable to discover the nonlinear mapping between LE and recombined images to synthesize the "virtual" recombined images. As a result, traditional FFDM can be enriched with the "virtual" recombined images. The objective of this study is to validate these two hypotheses by using a unique study procedure and two imaging datasets of both CEDM and FFDM images. The details of the study procedures and experimental results are reported in the following section of this paper.

## 2. MATERIALS

In this research, two separate datasets are used, which include a dataset acquired from tertiary medical center (Mayo Clinic Arizona), and a public dataset from INbreast (Moreira et al., 2012).

*2.1 Institutional Dataset from Mayo Clinic Arizona:*

Based on Institutional Review Board (IRB) approved study and data collection protocol, we reviewed CEDM examinations performed using the Hologic Imaging system (Bedford, MA, USA) between August 2014 and December 2015. All patients undertaken CEDM had a BI-RADS (Breast Imaging Reporting and Data Systems) (Liberman, L. and Menell, J.H., 2002) rating of 4 and 5 in their original FFDM screening images. Due to the detection of highly suspicious breast lesions, CEDM was offered as an adjunct test to biopsy in a clinical trial environment. All CEDM tests were performed prior to the biopsies. In summary, the patient cohort in this clinical trial had the following criteria: 1) the diagnostic mammogram was rated BI-RADS 4 or 5, and 2) histopathology test result was available from surgical or image-guided biopsy. We limited the cohort to BIRADS 4 and 5 lesions because the analysis required the gold standard of lesion pathology. 49 cases were identified that met the above inclusion criteria, which include 23 benign and 26 cancer biopsy-proven lesions. We analyzed one lesion per patient. If a patient had multiple enhancing lesions, the annotating radiologist used the largest lesion to ensure best feature selection. In CEDM, there are cranial-caudal (CC) and mediolateral-oblique (MLO) views for both LE and recombined images. Fig. 1 illustrates the example views on the LE and recombined images, respectively.

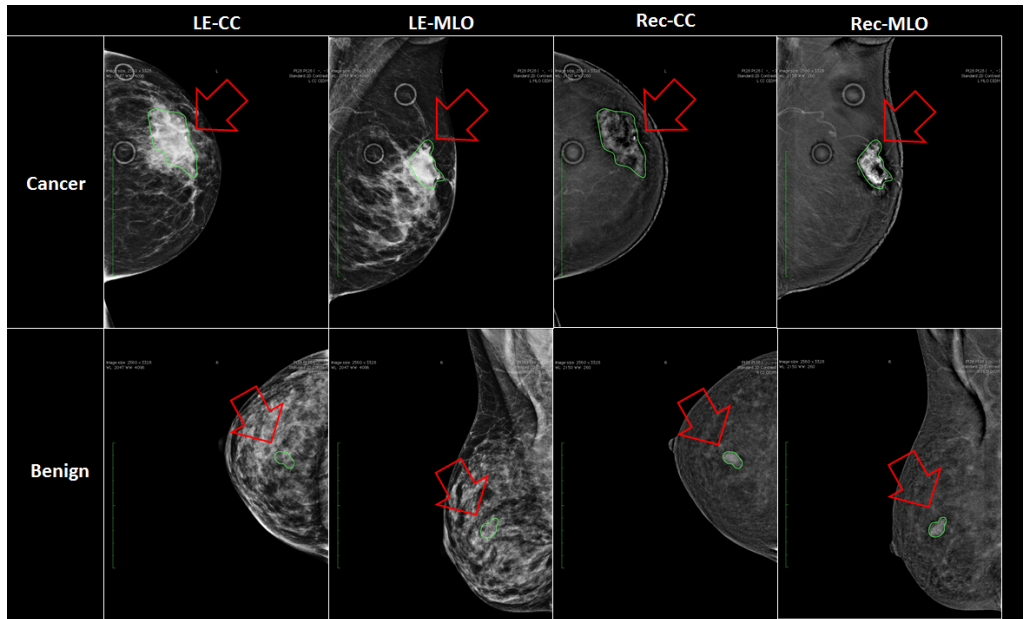

Fig. 1. Example of breast images (Cancer and Benign) for LE and recombined (Rec) images with 2 views (CC and MLO) (Lesions are highlighted with green circle).

For the 49 cases, all CEDM images with DICOM format were de-identified and transferred from the clinical PACS to a research database and loaded into the open source image processing tool OsiriX (OsiriX foundation, Geneva, Switzerland) (Rosset et al., 2004). DICOM images were anonymized and prepared for blinded reading by a radiologist. A fellowship trained breast radiologist with over 8 years of imaging experience interpreted the mammogram independently and used the OsiriX tool to outline lesion contours. Contours were drawn on recombined images (both CC and MLO views) for each patient on recombined images. These contours were then cloned onto LE images. All lesions were visible on both view CC and MLO views. This information is further used in the imaging pre-processing (see details in methodology section). Some examples LE and recombined images are shown in Fig. 1. As observed, LE images are not as easy as recombined images to visualize the lesions for both cancerous and benign cases.

*2.2 INbreast Public dataset:*

This dataset was obtained from INbreast, an online accessible full-field digital mammographic database (Moreira et al., 2012). INbreast was established by the researchers from the Breast Center in CHJKS, Porto, under the permission of both the Hospital's Ethics Committee and the National Committee of Data Protection. The FFDM images were acquired from the MammoNovation Siemens system with pixel size of 70 mm (microns), and 14-bit contrast resolution. For each subject, both CC and MLO view were available. For each image, the annotations of region of interests (ROIs) were made by a specialist in the field, and validated by a second specialist. The masks of ROIs were also made available. In this research, a dataset of 89 subjects was extracted by including subjects that have BI-RADS scores of 1, 2, 5 and 6. Subjects with BI-RADS 1 and 2 are regarded as benign tumor, and subjects with BI-RADS 5 and 6 are regarded as cancer. For each subject, images of CC and MLO view are used for feature extraction.

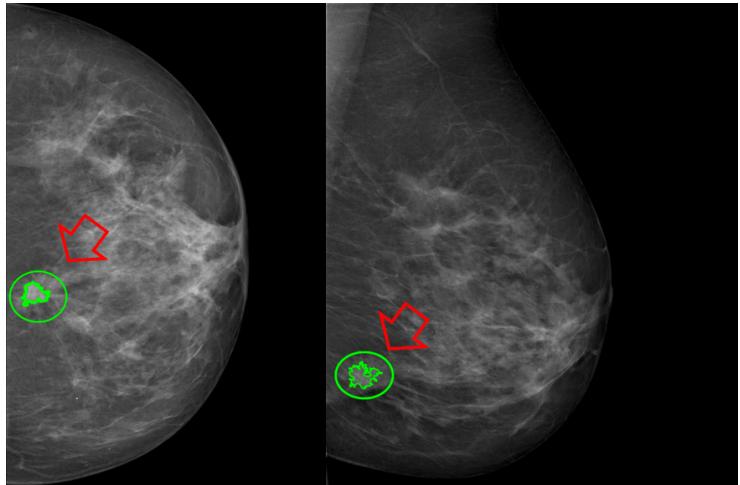

Fig. 2. Example of breast images for FFDM images from INbrease dataset with 2 views (CC on left and MLO on right) (Lesions are highlighted with green circle).

## 3 METHODOLOGY

To fully explore the advantages of CNNs and CEDM in breast cancer research, a Shallow-Deep CNN (SD-CNN) is proposed (Fig. 3). First, we develop a Shallow-CNN from CEDM to discover the relationships between LE images and recombined images. This Shallow-CNN is then applied to FFDM to render "virtual" recombined images. Together with FFDM, a trained Deep-CNN is introduced for feature extraction followed by classification models for diagnosis. Note for CEDM, we can start the workflow with the Deep-CNN directly.

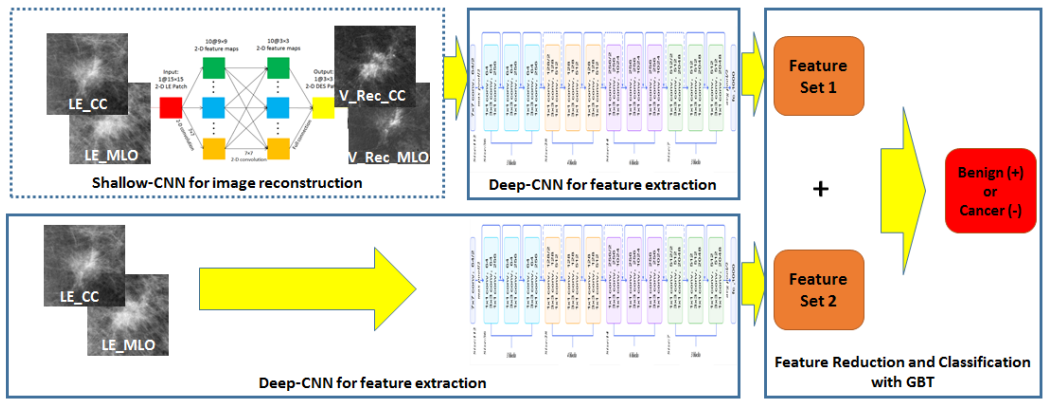

Fig. 3. Architecture of Shallow-Deep CNN

### 3.1 Image Pre-processing

Before the Deep-CNN and Shallow-CNN are employed, a four-step imaging pre-processing procedure is launched. First, for each image we identify a minimum-area bounding box that contains the tumor region. Specifically, for each tumor, we have a list of boundary points with coordinates in pair (x,y) available. The bounding box is decided using the $(x_{min}, y_{min})$ and $(x_{max}, y_{max})$ as the two diagonal corner points to ensure the box covers the whole tumor area. Note we have CC and MLO views for FFDM and we have CC and MLO views for both LE and recombined images for CEDM. As a result, there are two images from FFDM and four images from CEDM. The bounding box size varies case by case due to different sizes of tumors (ranging from 65×79 to 1490×2137 in this study). Next, an enlarged rectangle that is 1.44 times (1.2 times in width and 1.2 times in height) the size of bounding box is obtained. The enlarged bounding box approach is to include sufficient neighborhood information proved to increase the classification accuracy (Lévy and Jain, 2016). In the second step, this 'enlarged' rectangle is extracted and saved as one image. The third step is to normalize the image intensity to be between 0 and 1 using the max-min normalization. In the last step, the normalized images are resized to 224×224 to fully take advantage of trained ResNet model. Here we take the patches that contain tumor instead of the whole image as input. This is because the focus of the study is on tumor diagnosis and we believe the features generated by the deep-CNN from the tumor region shall better characterize the tumor, especially for the cases where the tumor region is small.

*3.2 Shallow-CNN: Virtual Image Rendering*

Inspired by the biological processes (Elmore et al., 2005), CNNs use a variation of multilayer perceptions designed to require minimal preprocessing. Individual neurons respond to stimuli only in a restricted region of the visual field known as the receptive field. This process is simulated through different layers (convolutional, pooling, fully connected). A CNN's capability is hidden behind the large amount trainable parameters which can be learned iteratively through gradient descent algorithms. In this research, a 4-layer CNN (Fig. 4) is implemented to model the latent relationship between the LE images (patches) and recombined images (patches). The model is then used to render "virtual" recombined images (patches) from FFDM images (patches).

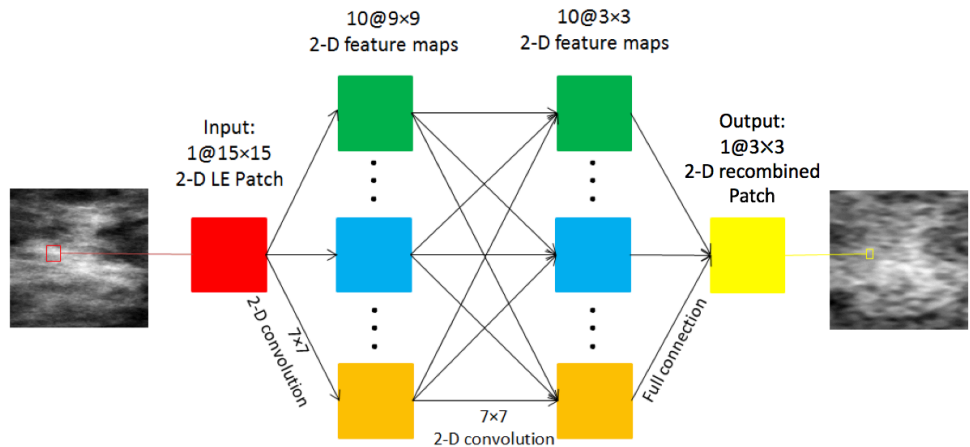

Fig. 4. Architecture of 4-layer shallow-CNN for "virtual" recombined image rendering

*3.3 Deep-CNN: Feature Generation*

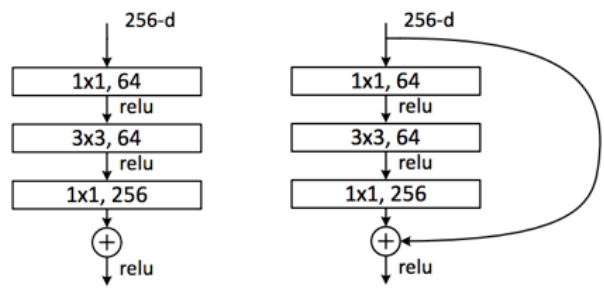

Fig. 5. Building blocks for traditional CNNs (left) and ResNet (right) (He et al., 2014)

ResNet is a trained deep CNN developed in 2015 with a revolutionary architecture using the "short-cut" concept in the building block. As seen in Fig. 5, the output of building blocks takes both final classification results and the initial inputs (the short-cut) when updating the parameters. As a result, it outperforms traditional deep-CNNs which are known to suffer from higher testing error since gradient tends to vanish as the number of layers increases (He et al., 2014). ResNet has different versions with 50, 101 and 152 layers but all based on the same building blocks. In the ImageNet competition, ResNet-50, ResNet-101 and ResNet-152 have comparable performances (top 5 error: 5.25% vs. 4.60% vs. 4.49%), but with quite different numbers of parameter (0.85M vs. 1.7M vs. 25.5M). For the consideration of balance between computation efficiency and accuracy, especially for the limited computation resources, we adopt ResNet-50 in this research.

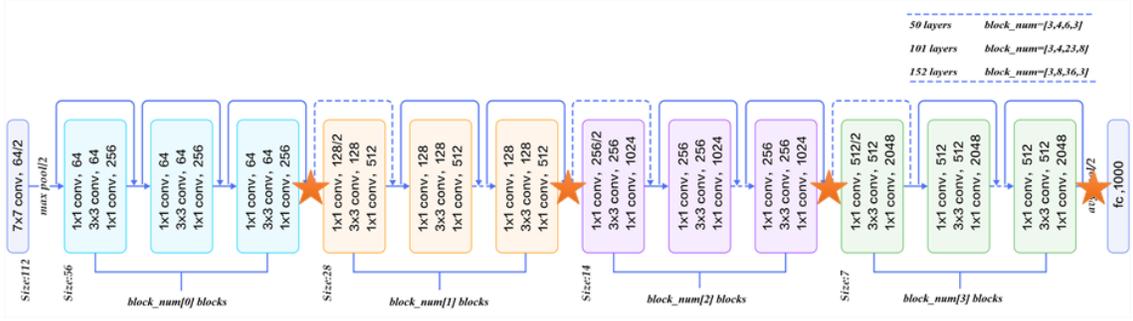

Fig. 6. Architecture of ResNet [37] (Red star are placed in layers where features are extracted; Dotted shortcuts increase feature dimensions by zero-padding; based on the output dimension of building blocks, the ResNet is divided into 4 different building blocks (BBs), they are shown with different colors in the figure (BB_1: blue (Fig. 5 Right), BB_2: orange, BB_3: purple, BB_4: green). Different version of ResNets vary in the number of BBs, for instance, the 50-layer version ResNet has 3 BB_1s, 4 BB_2s, 6 BB_3s and 3 BB_4s).

In general, ResNet consists of four types of buildings blocks. The CNN structures and the number of features for each block are shown in Fig. 6. We mark them with different colors. For simplicity, let blue for block type 1, orange for block type 2, purple for block type 3 and green for block type 4. ResNet-50 is defined as [3, 4, 6, 3] meaning that it has 3 type 1 blocks, 4 type 2 blocks, 6 type 3 blocks and 3 type 4 blocks. The output features are extracted from the final layer of each block type, that is, layer 10, 22, 40 and 49. Since we have no prior knowledge about the feature performance, we decide to take the features from all four layers (10, 22, 40 and 49) for the classification model development. For each feature map, the mean value is calculated and used to represent the whole feature map. The number of features extracted from each layer is listed in Table 1. For each view, we have 3840 (256+512+1024+2048) total features.

Table 1. Number of features from each layer for one image

| Layer # | 10 | 22 | 40 | 49 |
|---|---|---|---|---|
| # of features | 256 | 512 | 1024 | 2048 |

*3.4 Classification*

Boosting is a machine learning ensemble meta-algorithm aiming to reduce bias and variance (Bauer et al., 1999). It converts weak learners to strong ones by weighing each training sample inversely correlated to the performance of previous weak learners. Gradient boosting trees (GBT) is one of the most powerful boosting ensemble decision trees used in regression and classification tasks (Yang et al., 2017). It builds the model in a stage-wise fashion, and it generalizes them by allowing optimization of an arbitrary differentiable loss function. The nature of GBT makes it robust to overfitting by measuring the criterion it used when splitting the tree nodes. In addition, it provides the importance of each feature in the regression/classification for the ease of interpretation which is desirable in the medical applications. In GBT, the feature importance is related to the concept of Gini impurity (Rokach et akk., 2008). To compute Gini impurity ($I_G(p)$) for a set of items with $J$ classes, suppose $i \in \{1,2,\ldots,J\}$, and let $p_i$ be the fraction of items labeled with $i$ class in the set, we have:

$$I_G(p) = 1 - \sum_{i=1}^{J} p_i^2$$

When constructing each decision tree in the boosting classifier, a feature is used to divide the parent node into two children nodes based on a threshold. Since the decision tree is constructed with the goal being to minimize the overall Gini impurity, the post-splitting Gini impurity shall be smaller than the pre-splitting Gini impurity. The reduced Gini impurity thus can be used to as a measure of the contribution from the feature in the process of splitting the tree. The training procedure is to identify the optimal splitting features that offer the maximum impurity reduction (Yang et al., 2017) among the whole feature set. The process of building trees serves as feature selection and classification.

## 4 EXPERIMENTS AND RESULTS

The overall objective of this research is to demonstrate the clinical utility of our novel SD-CNN approach for breast cancer diagnosis. Therefore, we conduct two sets of experiments. The first experiment is to validate the values from recombined images for improved breast cancer diagnosis. Deep CNN, ResNet is applied. The second experiment is to investigate the feasibility of applying SD-CNN to enrich the traditional FFDM for improved diagnosis. A public FFDM dataset from INbreast is used and the results are compared with six state-of-the-art algorithms.

*4.1 Experiment I: Validating the Improved Accuracy in Breast Cancer Diagnosis on CEDM using Deep-CNN*

The workflow of our first experiment is shown in Fig. 7. Using 49 CEDM cases collected from Mayo Clinic Arizona, we first conduct the experiments using LE from CEDM images. For each subject in the dataset, LE images (both CC and MLO views) are processed through pre-processing procedure described in Section 3.1, after which 2 patches (224×224) are extracted. They are fed into the trained ResNet. As features from different layers of ResNet describe the image from different scales and aspects, in this research, we have all the features fed into the GBT to classify the case as cancer vs. benign. The procedures are implemented with a python library named "sklearn". Different settings to prevent the model from overfitting are used. For example, we set maximum depth of individual tree to be 3, use early stopping strategy by setting number of decision tree to be 21, max number of features to be searched for each split is $\sqrt{N}$ ($N$ is the number of features), the minimal number of samples falling in each leaf node is 2. Other settings are set to be default.

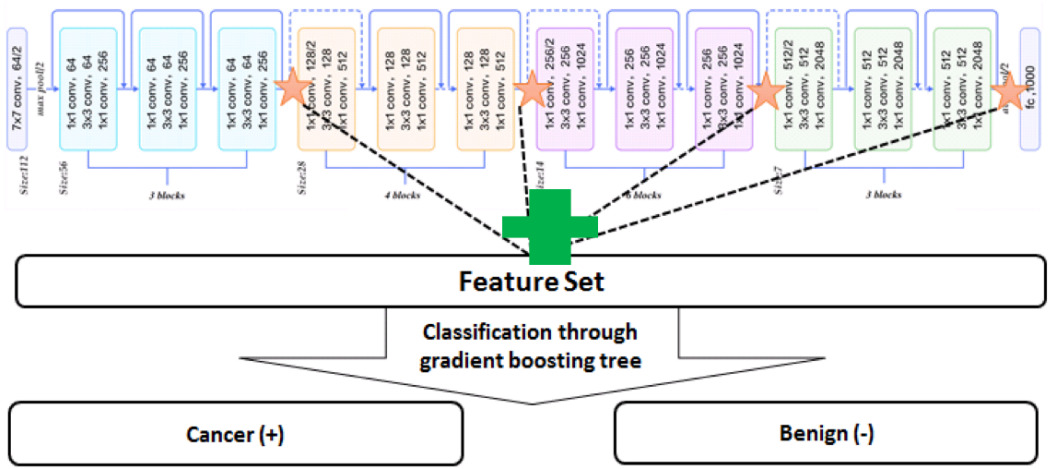

Fig. 7. Workflow of Experiment I

Next, we study the added values from recombined images for improved diagnosis. Specifically, CC and MLO view from recombined images are fed into the same pre-processing and feature generating procedure. The combination of LE and recombined image features are used in the classification model. Performance is measured based on leave-one-out cross validation to fully use the training dataset which is limited in size. Performance metrics are accuracy, sensitivity and specificity, and area under receiver operating characteristic curve(AUC) (see Table 2). The ROC curves for two models are shown in Fig.8. By using all the LE features generated by ResNet, we obtain the accuracy of 0.85 (Sensitivity=0.89 Specificity=0.80) and 0.84 for AUC. With additional features from recombined image, the model accuracy is improved to 0.89 (Sensitivity=0.93 Specificity=0.86) and AUC to 0.91.

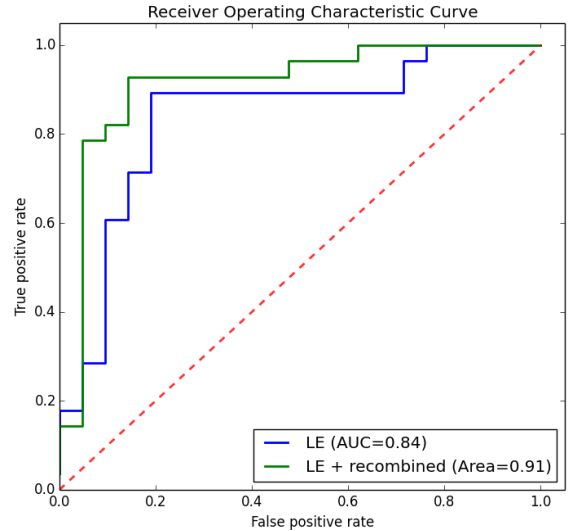

Fig. 8. Receiver operating characteristic curve for the model using FFDM image only vs. FFDM and recombined image

Table 2. Classification Performance of Experiment Using LE Images vs. LE and Recombined Images

|  | LE images | LE and Recombined images |
|---|---|---|
| Accuracy | 0.85 | 0.89 |
| Sensitivity | 0.89 | 0.93 |
| Specificity | 0.80 | 0.86 |
| AUC | 0.84 | 0.91 |

To explore the features contributing to the classification model, we calculated the contribution of each feature, and track the source image for each feature. The feature's importance score is measured through calculating the total impurity reduction when building the ensemble trees. (Note that the feature importance is calculated inside each leave-one-out loop, and the final result is the average for each feature among the loop). Table 3 summarizes the importance scores for the features from different sources (LE vs. Recombined Image). Here the scores are normalized by dividing individual score with summation of all scores. From Table 3, we observe among all the 99 features used in the model, 56 are from LE images which contribute 76.84% of the impurity reduction, 43 features are from recombined images which contribute to 23.16% in the modeling. The features from the recombined images help improve the accuracy of breast cancer diagnosis from 0.85 to 0.89.

Table 3. Contribution of features from different image sources

| Image Source | Number of features | Contribution of impurity reduction |
|---|---|---|
| LE image | 56 | 76.84% |
| Recombined image | 43 | 23.16% |

*4.2 Experiment II: Validating the Value of "Virtual" Recombined Imaging in Breast Cancer Diagnosis on FFDM Using SD-CNN*

The improved performance by adding the features from recombined images motivates us to study the validity of constructing and using the "virtual" recombined images from FFDM images for breast cancer diagnosis.

Here we first develop a 4-layer shallow CNN that learns the nonlinear mapping between the LE and recombined images using the same 49 CEDM dataset. CC and MLO view images are regarded as separate training data, so a total of

98 images are used, in which 5 subjects (10 images) are selected as validation material, and the rest 44 subjects (88 images) are sued as training material. By randomly extracting 2500 pair of training samples within masked tumor from each LE (input) and recombined image (output), a training dataset of 220000 (88×2500) samples is generated. The input samples for the CNN are 15×15 patches from LE images, the same input size as in (Li et al., 2016). Considering the relatively small receptive field and complexity of a shallow CNN, we set the output samples size as 3×3, it is our intention to explore the impact of the different output patch size for the breast cancer diagnosis as one of our future tasks. The output patches from recombined image are centered in the same position as input patches from the LE image. The input and output samples are fed into the CNN framework implemented with package of "Keras". The CNN has 2 hidden layers, with 10 7*7 filters in each layer. There are 5K trainable parameters through backpropagation with mini-batch gradient decent algorithm to increase learning speed. Batch size is set to be 128. The learning rate is set to be 0.01, ReLu activation function is used in all layers except the output layer, where activation function is not used. Other parameters are set to be default by "Keras" package. Finally, with the trained CNN and patches extracted from available modality, we can construct a "virtual" recombined image by assembling predicted patches into a whole image.

We use mean squared error (MSE) to evaluate the similarity between the "virtual" recombined image and the true recombined image for the 10 images in validation dataset. MSE measures the pairwise squared difference in intensity as:

$$MSE = \frac{1}{N}\sum_{i=1}^{N}|TRecombined(i) - VRecombined(i)|^2$$

Where N is total number of pixel in the selected patches, TRecombined(i) and VRecombined (i) are the intensity values for the same position in patches from the true recombined image and corresponding virtual recombined image.

For the 10 validation images, the MSE is 0.031 (standard deviation is 0.021). For illustration purpose, we choose four samples to demonstrate the resulting "virtual" recombined images vs. the true recombined images (Fig. 9). As seen, the abstract features (e.g., shape) and some details of the tumor from true recombined images are restored by the "virtual" recombined images.

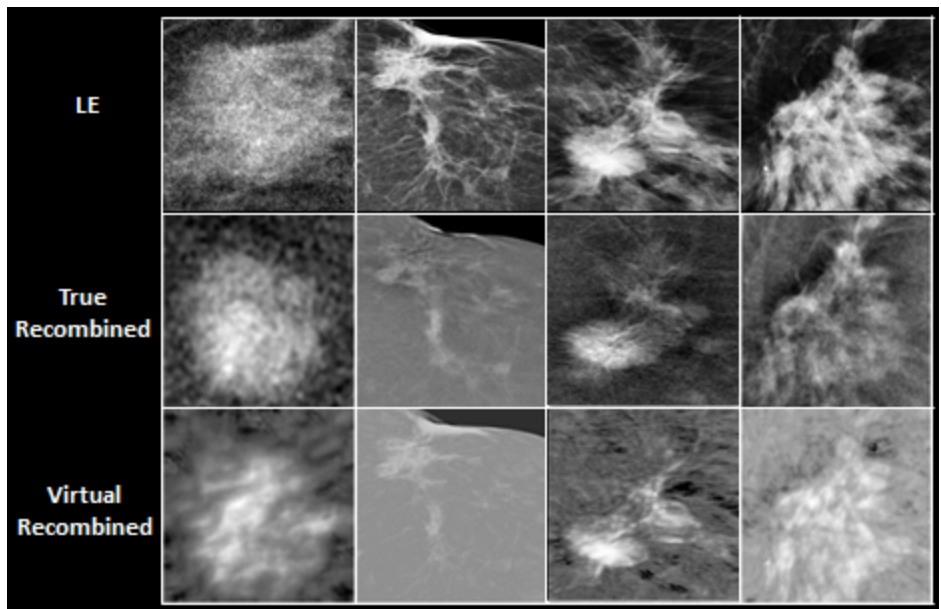

Fig. 9. Sample images of LE image, true recombined image and its corresponding virtual recombined in dataset I. (from left to right: benign, cancer, cancer, cancer)

With this trained shallow-CNN, we used the 89 FFDM cases from INbreast dataset to render the "virtual" recombined images. Specifically, for each subject, we slide the 15×15 window from left to right, top to bottom (step size = 1) in FFDM image, to get the input patches. The input patches are fed into the trained 4-layer CNN, from which we get the predicted virtual recombined image patches (3×3) as outputs. The small patches are placed at the same position as their corresponding input patches in the "virtual" recombined images. For the position with overlapping pixels, the values are replaced with mean value for all overlapping pixels. At last, the "virtual" recombined images are rendered. Fig. 10 illustrates some example FFDM images and their corresponding "virtual" recombined images. One clinical advantage of recombined image is it filters out dense tissues which often lead to false positive diagnosis. As seen from Fig. 10, the "virtual" recombined images preserve this advantage. Specifically, dense tissues surrounding tumors are excluded in "virtual" recombined images, making the core region easier to be identified (left two cases in Fig. 10). For the benign cases on the right, as the suspicious mass is mostly filtered out, it is mainly composed of dense tissues.

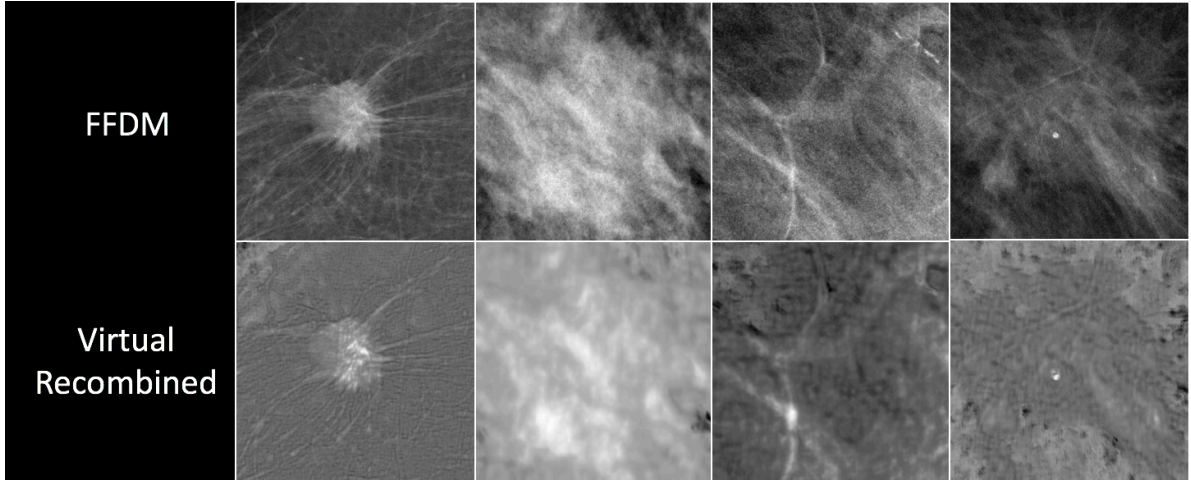

Fig. 10. Sample images of FFDM in dataset II and its corresponding "virtual" recombined (Two cases on left are cancerous with BI-RADS = 5, two cases on right are benign with BI-RADS=2)

Next, following the same procedure as the first experiment, we apply the ResNet on the FFDM alone, and on both FFDM and "virtual" recombined images together. ResNet is used for feature extraction followed by the GBT ensemble classifiers. The parameter for GBT settings is further tuned since the training dataset is slightly imbalanced (benign: cancer = 30: 59). The training weights for benign and cancer are set to be 1 and 0.5. Numbers of trees set to be 31. Other parameter settings remain the same as the first experiment and 10-fold cross validation is used. Fig. 11 shows the mean ROC curves for the model on FFDM alone vs. the model on FFDM and the "virtual" recombined image. The mean AUC for the classifier using FFDM features is $0.87 \pm 0.12$, while after adding the features from virtual recombined image, the AUC is increased to $0.92 \pm 0.14$. It is interesting to observe from Fig. 11 that sensitivities (true positive rate) of the two models have similar performance, the specificities (1 – false positive rate) vary greatly. We want to highlight the importance of specificity as breast cancer screening has high false positive recall rate (i.e., $\geqslant 10\%$). One known fact is that the probability that a woman will have at least one false positive diagnosis at 10 years screening program is 61.3% with annual and 41.6% with biennial screening (Michaelson et al., 2016). This will lead to additional MR exams (extra cost) and even biopsy. Another side effect is the negative psychological impacts. In this research, the use of recombined images ("virtual" recombined images) shows the great potential to address these challenges by improving the specificity. In Table 4, we summarize the model performances in terms of accuracy, sensitivity and specificity (threshold is set to be 0.75). While we observe that the model on FFDM vs. the model on FFDM and "virtual" recombined image show no significant differences on accuracy, sensitivity and even AUC, the performance on specificity shows significant improvements ($p<0.05$).

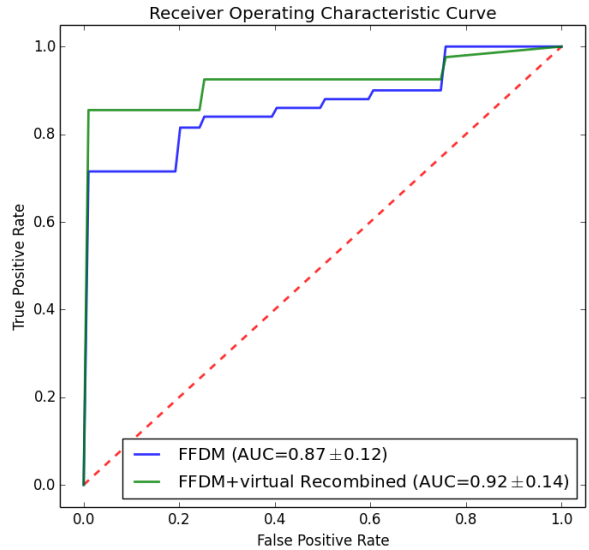

Fig. 11. Receiver operating characteristic curve for the model using FFDM image only verse FFDM and virtual recombined

Table 4. Classification Performance of Experiment Using FFDM Imaging vs. FFDM+Recombined imaging

|  | FFDM | FFDM + Virtual Recombined | P value |
|---|---|---|---|
| Accuracy | 0.84 ± 0.09 | 0.90 ± 0.06 | 0.14 |
| Sensitivity | 0.81 ± 0.16 | 0.83 ± 0.16 | 0.91 |
| Specificity | 0.85 ± 0.12 | 0.94 ± 0.04 | **< 0.05** |
| AUC | 0.87 ± 0.12 | 0.92 ± 0.14 | 0.28 |

In looking into the contributions from the features (Table 5), the use of "virtual" recombined imaging features improves the performances in terms of both accuracy and AUC. Calculation of contribution follows the same procedure as experiment I and is conducted inside each cross-validation loop. Among all the 154 features used in this experiment, 87 are from the "virtual" recombined image, which contribute 77.67% of the total impurity reduction. The rest 67 features are from LE images, and they contributed the rest 22.33% impurity reduction. It is interesting to observe from this experiment that the contributions from "virtual" recombined images are higher than the contributions from the true recombined images from the first experiment. One reason may be the second dataset has more denser tissue cases and it is believed recombined images shall be more useful in diagnosing the dense breast cases. This is yet to be confirmed with the radiologists which is our immediate next step.

Table 5. Contribution of features from different image sources

| Image Source | Number of features | Contribution of impurity reduction |
|---|---|---|
| LE image | 67 | 22.33% |
| Virtual Recombined image | 87 | 77.67% |

We further explore the state-of-the-art algorithms using the same INbreast dataset and compare our methods against the eight methods from the literature (Table 6). As seen, our approach using "virtual" recombined image outperforms six algorithms in terms of both accuracy and AUC. We want to highlight that one of papers by Dhungel et al. (2017) proposes four approaches. Among the four, the best performer has a 0.95 in accuracy and 0.91 in AUC, and the second performer has a 0.91 in accuracy and 0.87 in AUC. We conclude our approach has better AUC (0.92) comparing to both while inferior in accuracy (0.90). We contend that indeed, AUC is a more robust metric in the medical research and it is considered to be more consistent and have better discriminatory power comparing to accuracy (Huang et al. 2005).

Table 6 - Classification performance for using FFDM feature alone and using features from FFDM and "virtual" recombined and other state-of-the-art methods using INbreast dataset

| METHODS | ACC. | AUC |
|---|---|---|
| Random Forest on features from CNN with pre-training (Dhungel et al., 2017) | 0.95±0.05 | 0.91±0.12 |
| CNN + hand crafted features pre-training (Dhungel et al., 2017) | 0.91±0.06 | 0.87±0.06 |
| Random Forest + hand crafted features pre-training (Dhungel et al., 2017) | 0.90±0.02 | 0.80±0.15 |
| CNN without hand crafted features pre-training (Dhungel et al., 2017) | 0.72±0.16 | 0.82±0.07 |
| Multilayer perceptron (Sasikala, 2016) | 0.88 | 0.89 |
| Lib SVM (Diz et al., 2016) | 0.89 | 0.90 |
| Multi-kernel classifier (Augusto, 2014) | NA | 0.87 |
| Linear Discriminant analysis (Domingues et al., 2012) | 0.89 | NA |
| Our proposed approach on FFDM only | 0.84±0.09 | 0.87±0.12 |
| Our proposed approach on both FFDM and "virtual" Recombined Image | 0.90±0.06 | 0.92±0.14 |

## 5 DISCUSSION AND CONCLUSION

Differentiating benign cases from malignant lesions is one of the remaining challenges of breast cancer diagnosis. In this study, we propose a SD-CNN (Shallow-Deep CNN) to study the two-fold applicability of CNN to improve the breast cancer diagnosis. One contribution of this study is to investigate the advantages of recombined images from CEDM in helping the diagnosis of breast lesions using a Deep-CNN method. CEDM is a promising imaging modality providing information from standard FFDM combined with enhancement characteristics related to neoangiogenesis (similar to MRI). Based on our review of literature, no existing study has investigated the extent of CEDM imaging potentials using the deep-CNN. Using the state-of-art trained ResNet as a feature generator for classification modeling, our experiment shows the features from LE images can achieve accuracy of 0.85 and AUC of 0.84, adding the recombined imaging features, model performance improves to accuracy of 0.89 with AUC of 0.91.

Our second contribution lies in addressing the limited accessibility of CEDM and developing SD-CNN to improve the breast cancer diagnosis using FFDM in general. This the first study to develop a 4-layer shallow CNN to discover the nonlinear association between LE and recombined images from CEDM. The 4-layer shallow-CNN can be applied to render "virtual" recombined images from FFDM images to fully take advantage of the CEDM in improved breast cancer diagnosis. Our experiment on 89 FFDM dataset using the same trained ResNet achieves accuracy of 0.84 with AUC of 0.87. With the "virtual" recombined imaging features, the model performance is improved to accuracy of 0.90 with AUC of 0.92.

While promising, there is room for future work. First of all, the trained ResNet is a black-box feature generator, the features extracted may not be easy to be interpreted by the physicians. It is our intention to discover possible clinical interpretations from the features as one of our future tasks. For example, as the ResNet goes deeper, the initial layers of the features may represent the raw imaging characteristics as the first order statistics, the deeper layer of the features may represent the morphological characteristics (e.g., shape). This is yet to be explored. A second future work is related to the patch sizes. We plan to assess impacts of the different sized patches for both input and output images on the breast cancer diagnosis.